\documentclass[runningheads,a4paper]{llncs}
\usepackage[T1]{fontenc}
\usepackage{amssymb}
\usepackage{comment}
\usepackage{subcaption}
\usepackage{graphicx}
\usepackage{xspace}
\usepackage{booktabs}
\usepackage{tabularx}
\usepackage{listings}
\usepackage{color}
\usepackage{courier} 
\usepackage{amsmath}
\usepackage{stmaryrd}
\usepackage{xurl}
\usepackage[hidelinks]{hyperref}

\lstdefinelanguage{SPARQL}{
  morekeywords={SELECT,WHERE,FILTER,BIND,OPTIONAL,ORDER,BY,ASC,DESC,GROUP,HAVING,FUSEJOIN,SIMJOIN},
  sensitive=true,
  morecomment=[l]{\#},
  morestring=[b]"
}

\lstdefinestyle{sparqlstyle}{
  language=SPARQL,
  basicstyle=\ttfamily\footnotesize,   
  keywordstyle=\bfseries,             
  numbers=left,                      
  numberstyle=\tiny,
  numbersep=5pt,
  frame=single,                      
  columns=fullflexible,
  keepspaces=true,
  showstringspaces=false,
  breaklines=true,
  aboveskip=0.5\baselineskip,         
  belowskip=0.5\baselineskip,
  abovecaptionskip=2pt,                
  belowcaptionskip=0pt,
  captionpos=b                        
}

\begin{document}
\title{Scalable Uncertainty Reasoning in Knowledge Graphs}
%
%
\author{Jingcheng Wu\thanks{Category: Early Stage PhD}}
\authorrunning{J. Wu.}
%
\institute{University of Stuttgart, Stuttgart, Germany \\
\email{jingcheng.wu@ki.uni-stuttgart.de}}
\maketitle              
\begin{abstract}
Knowledge Graphs are pivotal for semantic data integration. The real-world data they model is often inherently uncertain. Within knowledge graphs, uncertainty manifests in three distinct levels: imprecise attribute values, probabilistic triple existence, and incomplete schema knowledge. However, current Semantic Web standards lack native support for reasoning over such uncertainty, and na\"ive extensions often incur computational intractability. In this thesis, I aim to develop a modular framework that addresses each level through tailored techniques: (1) defining probabilistic literals and a corresponding query algebra for continuous attributes; (2) a compilation-based framework transforming SPARQL provenance into tractable probabilistic circuits for uncertain triples; and (3) topology-aware geometric embeddings for statistical schema reasoning. The central hypothesis is that specialized reasoning mechanisms, namely algebraic, logical, and geometric approaches, can reconcile semantic precision with computational tractability.

\end{abstract}
\section{Introduction}
Knowledge Graphs (KGs)~\cite{DBLP:journals/csur/HoganBCdMGKGNNN21} serve as a versatile framework for the semantic integration of heterogeneous data~\cite{DBLP:journals/access/BuchgeherGGE21,lanza2024vision}, typically relying on W3C standards such as RDF~\cite{rdf11concepts} for data representation and SPARQL~\cite{sparql11query} for querying. While these standards excel at retrieving deterministic facts, they operate under an assumption of binary truth~\cite{DBLP:journals/ws/LukasiewiczS08}. For instance, the triple (:Motor123, rdf:type, :ElectricMotor) asserts that the entity Motor123 is classified as an electric motor. A standard query for this triple mandates an exact graph match, treating the statement as a binary fact that must explicitly exist in the dataset.

In reality, KGs are frequently incomplete and subject to uncertainty, containing missing links and unreliable facts that conflict with the deterministic nature of W3C standards~\cite{DBLP:conf/kdd/0001GHHLMSSZ14,DBLP:journals/ws/LukasiewiczS08,DBLP:conf/emnlp/ZhuWWZCKS25}. This pervasive uncertainty is not a monolithic phenomenon but in heterogeneous forms that defy a ``one-size-fits-all'' treatment. Consider the contrast between a triple representing a confirmed relationship with an uncertain value, such as (\texttt{:Motor123}, \texttt{:hasTemperature}, $\mathcal{N}(80^{\circ}\text{C}, 1^{\circ}\text{C})$), and a triple where the existence of the statement is a probabilistic hypothesis, such as (\texttt{:Grinder07812}, \texttt{:hasFault}, \texttt{:Overheat}) with $P = 0.12$. The former relies on calculus-based integration to evaluate range-based constraints over infinite continuous domains~\cite{DBLP:conf/dasfaa/ZhangCC10}. Conversely, the latter employs combinatorial model counting to aggregate the probabilities of discrete possible worlds~\cite{DBLP:conf/pods/Suciu20}.

Attempting to unify these heterogeneous forms by coupling discrete structural probabilities with continuous attribute distributions often incurs prohibitive computational complexity, such as \#P-hard inference~\cite{DBLP:journals/vldb/DalviS07}, because the engine cannot efficiently reconcile discrete counting with infinite continuous domains~\cite{DBLP:conf/aaai/SannerA12}. To achieve scalable reasoning without sacrificing semantic precision, we decompose this complexity by distinguishing the nature of the uncertainty. Specifically, following~\cite{DBLP:journals/tods/TaoXC07,DBLP:journals/kais/WangLLW13}, we categorize KG uncertainty into three distinct levels:

\begin{itemize}
    \item \textbf{Attribute-level Uncertainty} refers to the imprecision or inaccuracy of literal values~\cite{journal/fuzzyOnto,Compton2012SSN}. At this level, the existence of the triple is deterministic, but the literal value remains uncertain due to factors such as measurement errors or sensor noise. Recalling the aforementioned \textit{Motor123} example, the triple (\texttt{:Motor123}, \texttt{:hasTemperature}, $l$) is confirmed, but the literal $l$ follows the distribution $\mathcal{N}(80^{\circ}\text{C}, 1^{\circ}\text{C})$ rather than a precise scalar.
    \item \textbf{Triple-level Uncertainty} concerns the existential probability of a specific triple between two entities or an entity and a literal~\cite{DBLP:conf/kdd/0001GHHLMSSZ14,DBLP:journals/ws/LukasiewiczS08}. This uncertainty is frequently introduced by expert assessment~\cite{journal/fuzzyOnto}, link prediction~\cite{DBLP:conf/nips/BordesUGWY13,DBLP:conf/emnlp/DingWW0XT24}, or information extraction~\cite{DBLP:conf/kdd/0001GHHLMSSZ14}. Recalling the \texttt{Grinder07812} example, based on domain experts' observations, there is a 12\% probability of an overheat fault. Accordingly, the triple (\texttt{:Grinder07812}, \texttt{:hasFault}, \texttt{:Overheat}) is represented not as a binary fact, but as a probabilistic hypothesis where $P(\texttt{:Grinder07812}, \texttt{:hasFault}, \texttt{:Overheat}) = 0.12$.
    \item \textbf{Group-level Uncertainty} refers to the probabilistic constraints and statistical regularities residing within the KG schema~\cite{DBLP:conf/kr/LutzS10,DBLP:journals/corr/abs-2407-11821}. This level extends the scope of uncertainty from individual facts to terminological axioms regarding entire classes of entities. Instead of focusing on isolated instances, it models general dependencies. For example, regarding the class \texttt{:AngleGrinder} where $\texttt{:Grinder07812} \in \texttt{:AngleGrinder}$, statistical schema knowledge asserts that $85\%$ of such devices are equipped with a \texttt{:DustCover}. Formally, this is encoded as the probabilistic inclusion axiom $\texttt{:AngleGrinder} \sqsubseteq_{0.85} \exists \texttt{:hasPart}.\texttt{:DustCover}$.
\end{itemize}

Despite their differences, these three levels face a common barrier: existing Semantic Web engines lack native support for probabilistic reasoning. Attribute-level handling is currently restricted to descriptive metadata or inefficient sampling; Triple-level inference is \#P-hard without structural optimization; and Group-level logical reasoning scales poorly. This thesis addresses these limitations by extending the Semantic Web stack with distinct reasoning mechanisms tailored to each uncertainty type. Specifically, I study: (1) an algebraic query framework for Attribute-level uncertainty, instantiated via closed-form Gaussian Mixture Models (GMMs); (2) a logical compilation framework for Triple-level uncertainty, transforming probabilistic graph patterns into tractable circuits; and (3) a geometric embedding model for Group-level uncertainty, mapping statistical schemas onto topology-aware manifolds. The central hypothesis is that treating these levels through their respective algebraic, logical, and geometric lenses enables the reconciliation of semantic precision with computational tractability.

\section{State of the Art}
\label{sec:sota}
We analyze the state of the art based on the three levels.

\subsection{Attribute-level Uncertainty}
\label{sec:sota_attribute}
Existing standards within the Semantic Web, such as the Semantic Sensor Network (SSN) ontology~\cite{Compton2012SSN} and ProbOnto~\cite{DBLP:journals/bioinformatics/SwatGW16}, have established rich vocabularies for describing probability distributions and sensor observations. Similarly, SCOVO~\cite{DBLP:conf/esws/HausenblasHRFA09} and the RDF Data Cube~\cite{W3C:DataCube} provide ontologies for exchanging statistical data. However, these frameworks function primarily as \textit{descriptive metadata schemas}. They define how statistical data is structurally represented but lack the underlying query algebra required to execute operations, such as convolution or Bayesian fusion, directly within the database engine.

In the RDF stream processing, Keskisärkkä et al.~\cite{DBLP:conf/ekaw/KeskisarkkaBLH20,DBLP:conf/i-semantics/KeskisarkkaBH21} introduced a custom literal datatype and SPARQL extensions for probabilistic filtering within the RSP-QL* model. While sharing the high-level idea of embedding distributions in RDF literals, their work targets the transient nature of data streams and differs from our persistent KG framework in three key respects. First, their representation conflates random variables with distributions, whereas our framework separates them to uniformly handle multi-dimensional data. Second, their operations map distributions strictly to scalar probabilities without algebraic closure, precluding chained transformations (e.g., convolution, Bayesian fusion) or distributional comparisons like similarity joins. Third, they support only basic parametric families, whereas our polymorphic support for heterogeneous distribution families (including GMMs, Dirichlet, and histograms) moving beyond basic parametric forms.

Beyond the Semantic Web, handling continuous variables in probabilistic databases often incurs prohibitive computational costs. In relational databases, systems like Orion~\cite{DBLP:conf/comad/SinghMMPHS08} and MCDB~\cite{DBLP:conf/sigmod/JampaniXWPJH08} allow multiple probability distributions as tuple attributes. Orion defines algebraic operations such as floor, marginalize, and product over distributions, whereas MCDB relies on generating thousands of random samples during query evaluation. Although the Monte Carlo approach offers flexibility, it introduces massive runtime overhead, rendering it unsuitable for the interactive querying demands of large-scale knowledge graphs.

These limitations reveal a critical research gap: the absence of a query algebra that treats diverse probability distributions (encompassing discrete, continuous, parametric, and non-parametric forms) as \textit{first-class citizens} within the RDF data model. Grounded in standard random variable modeling, this algebra supports algebraically closed distribution-to-distribution transformations and similarity joins across heterogeneous families, while maintaining computational efficiency through a hybrid of closed-form operations and different sampling strategies~\cite{DBLP:books/sp/RobertC04,montecarlo,sequentialtest}.


\subsection{Triple-level Uncertainty}
\label{sec:sota_triple}



This level addresses the existential probability of relations between entities. The foundational framework is the Tuple-Independent Database (TID)~\cite{DBLP:journals/ftdb/BroeckS17,DBLP:conf/kr/CeylanDB16,DBLP:conf/pods/Suciu20}, where every triple is an independent Bernoulli event and query evaluation maps to Weighted Model Counting on the provenance. However, real-world triples frequently exhibit dependencies modeled via PGMs~\cite{DBLP:journals/vldb/LiSD11}, and integrating such structures into SPARQL evaluation remains open.


Even under TID, the Dichotomy Theorem~\cite{DBLP:journals/jacm/DalviS12} classifies conjunctive queries into Safe (PTIME) and Unsafe (\#P-hard) classes. Safe queries admit lifted inference directly from query structure, but existing provenance engines~\cite{DBLP:conf/www/AsmaHGFFH24,DBLP:journals/pvldb/HernandezGH21} based on the semiring framework fail to exploit this classification. While these systems provide a general algebraic approach for metadata propagation, they treat all query patterns uniformly, missing opportunities to perform efficient lifted inference for safe graph patterns.


For the full expressivity of SPARQL, including non-monotonic operators (OPTIONAL, MINUS), Geerts et al.~\cite{DBLP:journals/jacm/GeertsUKFC16} introduced spm-semirings to capture their provenance semantics, implemented by SPARQLProv~\cite{DBLP:journals/pvldb/HernandezGH21} and NPCS~\cite{DBLP:conf/www/AsmaHGFFH24}. However, probabilistic evaluation of the resulting lineage remains computationally prohibitive~\cite{DBLP:journals/ftdb/BroeckS17}.

To mitigate this computational intractability, the database community leverages knowledge compilation (KC)~\cite{DBLP:journals/corr/abs-1106-1819}. This paradigm shifts the complexity from online query evaluation to an offline compilation phase, transforming the lineage formula into a tractable target language, notably deterministic Decomposable Negation Normal Form (d-DNNF)~\cite{DBLP:journals/ftdb/BroeckS17,DBLP:conf/kr/CeylanDB16}. Unlike raw lineage expressions, d-DNNFs satisfy determinism and decomposability, two structural properties that together reduce inference complexity from \#P-hard to linear in the circuit size. While KC is highly effective for monotonic relational queries, adapting it to handle the non-monotonic semantics of full SPARQL with spm-semirings provenance remains a significant open problem that this thesis also aims to address.



We note that Fuzzy Logic~\cite{DBLP:journals/ws/LukasiewiczS08,DBLP:conf/cilc/Straccia12,DBLP:journals/ws/ZimmermannLPS12} assigns degrees of truth rather than probabilities and lacks the statistical compositionality required for rigorous inference. This thesis therefore focuses on probabilistic semantics.

These observations identify three interrelated gaps: (1) the failure to exploit query structure for lifted inference; (2) the absence of efficient inference mechanisms for non-monotonic provenance; and (3) the lack of native support for tuple dependencies beyond the TID assumption. This thesis addresses these gaps through a unified compilation framework that transforms SPARQL lineage into tractable circuit representations.

\subsection{Group-level Uncertainty}
\label{sec:sota_group}

This level addresses statistical constraints over terminological knowledge, asserting conditional probabilities such as ``$85\%$ of angle grinders are equipped with a dust cover.'' The foundational formalism for this reasoning is Statistical $\mathcal{EL}$ (SEL)~\cite{DBLP:conf/sum/PenalozaP17}. While SEL provides rigorous semantics for these probabilistic terminological axioms, theoretical analyses establish that its exact satisfiability checking is \textsc{Exptime}-complete~\cite{DBLP:journals/ipl/Bednarczyk21}. This high complexity renders classical inference mechanisms relying on linear programming~\cite{DBLP:conf/kr/LutzS10} or tableau algorithms~\cite{DBLP:conf/jelia/BothaMP19} computationally infeasible for large-scale KGs~\cite{DBLP:journals/corr/abs-2407-11821}.

To overcome the scalability barrier, recent research has pivoted towards geometric neuro-symbolic approximations~\cite{DBLP:conf/www/GregucciN0S23,10.1007/978-3-031-19433-7_2}. Representative geometric embedding models, exemplified by BoxEL-based methods, map concepts to axis-parallel boxes in vector space~\cite{10.1007/978-3-031-19433-7_2}. This paradigm shift replaces intractable logical deduction with polynomial-time geometric measurement, modeling the conditional probability $P(D \mid C)$ as a volumetric ratio~\cite{DBLP:journals/corr/abs-2407-11821}:
\begin{equation}
    P(D \mid C) \approx \frac{\text{Vol}(\text{Box}(C) \cap \text{Box}(D))}{\text{Vol}(\text{Box}(C))}
\end{equation}
However, these models typically operate within flat Euclidean space, creating a fundamental topological mismatch when representing intrinsically hierarchical ontologies such as taxonomies~\cite{DBLP:conf/kdd/XiongZNXP0S22}. Euclidean embeddings often require high dimensionality to preserve such structures, leading to parameter inefficiency and systematic approximation errors~\cite{DBLP:conf/nips/NickelK17}.

This observation identifies a critical research gap: the absence of topology-aware geometric embeddings that match the structural characteristics of ontologies to appropriate manifolds, thereby improving approximation fidelity while maintaining computational tractability.

\section{Problem Statement and Contributions}
\label{sec:prob}

In this thesis, I will investigate: 1) how to extend SPARQL to support algebraic operations over continuous probability distributions for attribute-level uncertainty; 2) how to efficiently evaluate complex, non-monotonic SPARQL queries by compiling their lineage into tractable probabilistic circuits~\cite{DBLP:journals/ftdb/BroeckS17,DBLP:journals/ai/CeylanDB21} for triple-level uncertainty; and 3) how to approximate intractable terminological reasoning by embedding concepts into topology-aware geometric manifolds (e.g., Hyperbolic space) for group-level uncertainty.

\medskip
\noindent \textbf{Problem 1: Attribute-level Uncertainty.} An RDF graph $\mathcal{G}$ is a finite set of triples $(s, p, o) \in (\mathcal{I} \cup \mathcal{B}) \times \mathcal{I} \times (\mathcal{I} \cup \mathcal{B} \cup \mathcal{L})$, where $\mathcal{I}, \mathcal{B}$, and $\mathcal{L}$ denote disjoint sets of IRIs, blank nodes, and literals, respectively. The evaluation of a SPARQL query $Q$ over $\mathcal{G}$, denoted as $\llbracket Q \rrbracket_\mathcal{G}$, relies on deterministic pattern matching to yield a set of solution mappings $\mu: V \to (\mathcal{I} \cup \mathcal{B} \cup \mathcal{L})$. A fundamental representational gap exists because the literal set $\mathcal{L}$ consists strictly of deterministic values. Standard RDF specifications provide no built-in datatype to encode continuous random variables, such as the multi-dimensional distributions inherently present in sensor measurements. Furthermore, the standard SPARQL algebra lacks built-in operators to manipulate such probabilistic data. Specifically, it offers no formal semantics to evaluate probabilistic filter constraints (e.g., \texttt{FILTER(P(?X > ?c) >= ?theta)}) or to execute closed-form algebraic transformations, such as distribution fusion, as native query operations.

\medskip
\noindent \textbf{Hypothesis 1.} Extending SPARQL with a native datatype for continuous random variables and corresponding closed-form algebraic operators, instantiated via GMMs (universal approximators admitting closed-form solutions for operations such as convolution and product), allows for the manipulation of uncertain data with semantic precision comparable to Monte Carlo simulations but at significantly lower latency.

\medskip
\noindent The related research questions are:

\begin{itemize}
    \item \textbf{RQ 1.1.} How can multi-dimensional sensor measurement data be efficiently represented as random variables and treated as first-class citizens within the RDF data model?

    \item \textbf{RQ 1.2.} How can common probabilistic functions be executed as SPARQL operations that interact seamlessly with other knowledge graph representations?

    \item \textbf{RQ 1.3.} How does the proposed algebraic approach compare to sampling-based baselines in terms of query execution time and approximation error?
\end{itemize}

\noindent \textbf{Problem 2: Triple-level Uncertainty.} We address the challenge of scalable computation of exact probabilities for SPARQL queries over a probabilistic knowledge graph $\mathcal{G}_{prob}$. Formally, we define $\mathcal{G}_{prob}$ as a pair $(\mathcal{G}, \mathcal{P})$, where $\mathcal{G}$ is an RDF graph and $\mathcal{P}$ denotes a joint probability distribution over the subgraphs of $\mathcal{G}$. The TID model parameterizes the joint distribution by associating a marginal probability $P(t) \in (0, 1]$ with each triple $t \in \mathcal{G}$, assuming independence such that $\mathcal{P}$ is determined by the factorization $\mathcal{P}(W) = \prod_{t \in W} P(t) \prod_{t \in \mathcal{G} \setminus W} (1 - P(t))$ for any world $W \subseteq \mathcal{G}$. Crucially, accurately modeling real-world data requires extending beyond the TID assumption to capture complex dependencies between triples, thereby incorporating conditional probabilities to specify $\mathcal{P}$. The query semantics are grounded in the possible worlds framework, where the sample space $\Omega$ consists of $2^{|\mathcal{G}|}$ deterministic subgraphs of $\mathcal{G}$, each weighted by $\mathcal{P}$.

As discussed in Section \ref{sec:sota_triple}, three barriers remain. First, existing engines do not exploit the Safe/Unsafe classification to perform lifted inference for tractable query classes. Second, while spm-semirings provide a representational solution for non-monotonic operators, no efficient inference mechanism exists to evaluate the resulting lineage expressions. Third, integrating probabilistic dependencies modeled via PGMs with SPARQL query evaluation remains an open challenge.

\medskip
\noindent \textbf{Hypothesis 2.} Compiling provenance lineage and probabilistic dependencies into probabilistic circuits (e.g., d-DNNF~\cite{DBLP:journals/jair/DarwicheM02}) enables scalable exact inference for SPARQL queries, reducing the inference complexity from exponential to polynomial in the size of the circuit.

\medskip
\noindent The related research questions are:

\begin{itemize}
    \item \textbf{RQ 2.1.} How can independent sub-patterns within a SPARQL query plan be identified and exploited during knowledge compilation to achieve lifted inference for monotonic fragments?
    \item \textbf{RQ 2.2.} How can lineage polynomials defined over spm-semirings be correctly compiled into probabilistic circuits while preserving the probabilistic semantics of non-monotonic operators?
    \item \textbf{RQ 2.3.} How can probabilistic dependencies between triples, modeled via Bayesian Networks, be efficiently compiled into a circuit representation and integrated with the query lineage to support exact inference over correlated data?
\end{itemize}

\noindent \textbf{Problem 3: Group-level Uncertainty.} This level addresses the integration of statistical terminological knowledge into knowledge graph querying. Formally, we consider Statistical $\mathcal{EL}$ ($\mathcal{SEL}$), a probabilistic extension of the lightweight Description Logic $\mathcal{EL}$ where probabilities represent proportions in the domain. A $\mathcal{SEL}$ knowledge base is a tuple $\mathcal{K} = (\mathcal{T}, \mathcal{A})$, consisting of a TBox $\mathcal{T}$ and an ABox $\mathcal{A}$. The TBox $\mathcal{T}$ contains both classical axioms (e.g., $C \sqsubseteq D$) and statistical conditionals of the form $(D \mid C)[p]$, asserting that the proportion of $C$-instances that are also $D$-instances is $p$. Classical subsumption $C \sqsubseteq D$ arises as the special case where $p = 1$. The ABox $\mathcal{A}$ contains assertions about specific individuals. Given an ABox $\mathcal{A}$ and a target concept $C$, the reasoning task is to compute the probability of instance membership $a : C$, which requires projecting the statistical proportions from $\mathcal{T}$ onto the individual $a$ conditioned on the explicit assertions in $\mathcal{A}$.

Suppose $\mathcal{T}$ contains the conditional $(\texttt{HasFault} \mid \texttt{AngleGrinder})[0.12]$, and the ABox asserts $\texttt{Grinder07812} : \texttt{AngleGrinder}$ without explicit fault information. A probabilistic query asking for all potentially faulty devices should return \texttt{Grinder07812} with probability $0.12$, derived from the statistical schema constraint. This reasoning pattern, propagating schema-level statistics to instance-level probability estimates, constitutes the core computational task of group-level uncertainty.

As discussed in Section \ref{sec:sota_group}, two fundamental challenges impede this reasoning task. First, exact probabilistic inference in $\mathcal{SEL}$ is \textsc{Exptime}-complete, motivating the use of approximate inference methods to achieve scalability. Second, while geometric embedding approaches such as BoxEL~\cite{10.1007/978-3-031-19433-7_2} offer tractable approximations, their representational capacity is constrained by the geometry of flat Euclidean space, which fails to preserve hierarchical structures without incurring excessive dimensionality.

\medskip
\noindent \textbf{Hypothesis 3.} Mapping ontological concepts to manifolds that match their structural characteristics (e.g., hyperbolic space for hierarchies) improves approximation fidelity compared to flat Euclidean embeddings, while maintaining polynomial-time inference complexity.

\newpage
\medskip
\noindent The related research questions are:

\begin{itemize}
    \item \textbf{RQ 3.1.} How can probabilistic box embeddings be generalized from Euclidean space to non-Euclidean manifolds (e.g., Hyperbolic space) while preserving tractable volume computation for conditional probability estimation?
    \item \textbf{RQ 3.2.} How can the structural characteristics of an ontology (e.g., hierarchical depth, branching factor) guide the selection of appropriate geometric spaces to improve approximation fidelity?
\end{itemize}

\section{Research Methodology and Approach}

In this section, I describe the proposed approach to address the problems identified in Section \ref{sec:prob}.

\subsection{Approach to Attribute-level Uncertainty}

To address the limitations identified in Section~\ref{sec:sota_attribute}, my approach embeds probability distributions as first-class citizens within the RDF data model, instantiating the framework using GMMs, which serve as universal approximators that admit closed-form algebraic solutions.

To address \textbf{RQ 1.1} and \textbf{RQ 1.2}, I propose a two-layered extension to the Semantic Web stack. At the data level, I will define a custom RDF datatype to represent continuous random variables by encoding their probability distributions as structured literals. At the query level, I will extend the SPARQL algebra with distribution-aware operators, including probabilistic filtering ($P(X > c) \geq \theta$), Bayesian fusion, and similarity joins. This design ensures that uncertainty reasoning is handled natively within the query algebra rather than through external application logic. Addressing \textbf{RQ 1.3}, I will benchmark the algebraic framework against Monte Carlo baselines to quantify the trade-off between query latency and approximation fidelity. Furthermore, I will specifically evaluate the performance impact of the proposed extensions by comparing the execution times of queries utilizing similarity joins and filters against standard SPARQL benchmarks.


To evaluate these contributions, I construct synthetic knowledge graphs scaling up to 3 million triples with varying distribution complexities (e.g., mixture models with $K \in \{1, 3, 5, 10\}$ components). Since standard SPARQL benchmarks lack uncertainty, I generate parallel deterministic datasets (replacing each GMM with its scalar mean) to isolate the overhead of probabilistic extensions. Beyond overhead ratio, mean absolute error, and latency scaling, the evaluation targets two additional dimensions. First, I compare the dedicated \texttt{SIMJOIN} operator against a na\"ive \texttt{BIND}+\texttt{FILTER} pipeline to quantify the benefit of distribution-aware pruning via the Data Processing Inequality~\cite{DBLP:journals/qic/BeaudryR12}. Second, I evaluate multiple sampling strategies (na\"ive Monte Carlo, stratified sampling~\cite{DBLP:books/sp/RobertC04}, sequential testing~\cite{sequentialtest}, and an adaptive cascade) to assess per-pair latency and classification accuracy across varying difficulty levels.

\subsection{Approach to Triple-level Uncertainty}

To address the limitations identified in Section~\ref{sec:sota_triple}, my approach leverages the framework of probabilistic circuits~\cite{ProbCirc20}. This technique shifts the computational burden from online query evaluation to an offline compilation phase, producing tractable circuit representations that enable linear-time inference. To integrate this technique into the Semantic Web stack, the system operates as a query-rewriting middleware. It intercepts standard SPARQL queries, evaluates them under possible-worlds semantics, and returns solution mappings annotated with marginal probabilities.

To tackle \textbf{RQ 2.1}, a static analysis phase classifies queries according to the Safe/Unsafe dichotomy~\cite{DBLP:journals/jacm/DalviS12}. For Safe queries, the system rewrites the query into a safe evaluation plan that decomposes probability computation along independent sub-queries, enabling lifted inference without compilation. For Unsafe queries, the provenance lineage is compiled into d-DNNF circuits. To study \textbf{RQ 2.2}, this compilation must correctly handle the monus operator ($\ominus$) from spm-semiring provenance while preserving decomposability for tractable inference. Answering \textbf{RQ 2.3}, I will extend the framework to incorporate probabilistic dependencies modeled as Bayesian Networks. By encoding the network structure as Conjunctive Normal Form (CNF) constraints and compiling them jointly with the query lineage, the system captures both query structure and tuple correlations in a unified circuit representation.

To mitigate the potential exponential cost of compilation, I plan to investigate circuit caching techniques to amortize the offline overhead for workloads with recurring query patterns.

For evaluation, correctness is verified by comparing circuit-based inference against brute-force enumeration over all $2^{|G|}$ possible worlds on small-scale probabilistic KGs ($|G| \leq 20$). Efficiency is assessed by measuring compilation time, circuit size, and online inference time as functions of graph size and query complexity. I additionally compare lifted inference on Safe queries against full compilation to quantify the benefit of exploiting query structure.

\subsection{Approach to Group-level Uncertainty}

To address the limitations identified in Section~\ref{sec:sota_group}, my approach adopts the geometric approximation paradigm, modeling ontological axioms as spatial containment in the embedding space and estimating conditional probabilities via volumetric ratios.

Existing models such as BoxEL~\cite{10.1007/978-3-031-19433-7_2} operate in flat Euclidean space, where representing deep hierarchies requires high-dimensional embeddings because Euclidean volume grows only polynomially with the radius. To address \textbf{RQ 3.1}, I plan to extend box embeddings to non-Euclidean manifolds, such as hyperbolic space, where volume grows exponentially with the radius, naturally matching the exponential branching of taxonomic hierarchies while requiring fewer dimensions. Addressing \textbf{RQ 3.2}, I will investigate how structural properties of an ontology, such as Gromov hyperbolicity, correlate with embedding quality. This study aims to provide theoretical guidance on the conditions under which non-Euclidean embeddings offer clear advantages.

Approximation fidelity is evaluated against exact inference on small-scale $\mathcal{EL}$ knowledge bases using mean absolute error. I compare Euclidean BoxEL against hyperbolic embeddings at equal dimensionality, varying ontology depth and branching factor to test Hypothesis~3. Scalability is assessed by measuring training and inference time as a function of ontology size.

\section{Preliminary Results}
\label{sec:results}

To natively integrate probabilistic literals into SPARQL for attribute-level uncertainty, I developed \texttt{ProbSPARQL}. It ensures algebraic closure through distribution-valued operators (e.g., convolution) while extracting scalar statistics via numeric ones (e.g., \texttt{CDF}, \texttt{JSD}). This framework, which additionally introduces a divergence-based similarity join (\texttt{SIMJOIN}), is implemented atop Apache Jena/Fuseki.

Benchmarks on 3M-triple KGs show a $1.25\times$--$2.16\times$ constant overhead over deterministic baselines. Pushing probabilistic filters below joins yields up to a $17.7\times$ speedup. For \texttt{SIMJOIN}, Data Processing Inequality pruning eliminates $90\%$ of candidates, achieving $178.8\times$ end-to-end speedup. The architecture supports histogram and Dirichlet distributions. Histogram-based similarity computation is three orders of magnitude faster than Monte Carlo sampling. Code and datasets are available.\footnote{\url{https://github.com/0sidewalkenforcer0/ProbSPARQL}}

\section{Conclusions}


This thesis addresses scalable uncertainty reasoning in KGs through a decompose-then-specialize methodology: algebraic operators for attribute-level distributions, compilation-based circuits for triple-level provenance, and geometric embeddings for group-level statistical schemas. The preliminary results on \texttt{ProbSPARQL} (Section~\ref{sec:results}) validate this strategy for the attribute level. While restricting group-level reasoning to $\mathcal{EL}$ limits expressivity, $\mathcal{EL}$ underpins major ontologies (e.g., SNOMED CT~\cite{schulz2009snomed}, Gene Ontology~\cite{ashburner2000gene}) and provides a natural starting point. Extensions to more expressive logics remain an open question.

\section*{Acknowledgments}

I would like to express sincere gratitude to my supervisor, Prof. Dr. Steffen Staab, and my co-supervisors, Dr. Ratan Bahadur Thapa and Dr. Daniel Hern\'{a}ndez, for their continuous support and guidance throughout this research. This Ph.D. project is funded by the Deutsche Forschungsgemeinschaft (DFG, German Research Foundation) within the Collaborative Research Center SFB 1574, Project number 471687386.

\newpage
\bibliographystyle{splncs04}
\bibliography{ref}
\end{document}